\documentclass[preprint,12pt]{elsarticle}

\usepackage{amssymb}
\usepackage{xcolor}
\usepackage{amsmath}

\journal{Journal of Computational Science}

\begin{document}

\begin{frontmatter}



\title{Decision-Driven Geosteering Under Uncertainty: A Unified Framework for Sequential Decision Optimization}


\author[label1,label2]{Hibat Errahmen Djecta\corref{cor1}}
\ead{hidj@norceresearch.no}
\author[label1]{Sergey Alyaev}
\author[label1]{Kristian Fossum}
\author[label2]{Reidar B. Bratvold}
\author[label2]{Ressi Bonti Muhammad}
\author[label3]{Apoorv Srivastava}

\cortext[cor1]{Corresponding author}

\affiliation[label1]{
  organization={NORCE Research Centre},
  city={Bergen},
  country={Norway}
}

\affiliation[label2]{
  organization={University of Stavanger},
  city={Stavanger},
  country={Norway}
}

\affiliation[label3]{
  organization={Stanford University},
  city={Stanford},
  state={CA},
  country={USA}
}

\begin{abstract}
Geosteering requires navigating a well trajectory through an unknown geological configuration, while sequentially updating decisions based on indirect measurements acquired during drilling. This work presents an uncertainty-aware geosteering framework that tightly integrates particle filtering for probabilistic subsurface interpretation with value-based reinforcement learning for sequential decision-making. Geological uncertainty ahead of the drill bit is represented explicitly through a particle filter (PF), enabling belief-informed control rather than deterministic trajectory correction.

The framework couples PF belief updates with belief-informed decision policies and evaluates three decision-making options that operate under identical uncertainty representations: an interpretable Approximate Dynamic Programming (ADP) scheme, a Deep Q-learning baseline, and a Dual Deep Reinforcement Learning (Dual DRL) architecture trained with a target Q-network scheme for stability, using a dueling (value/advantage) decomposition for Q-value parameterization. Beyond final placement performance, we assess policy behavior using stability-oriented metrics that quantify steering smoothness over time, providing additional operational insight into how decision policies respond as uncertainty evolves.

The framework is integrated with an API for validation within an industrial geosteering simulator under realistic measurement noise and drilling constraints. Using identical geological realizations, operational limits, and reward definitions across methods, the experiments provide a controlled and high-fidelity evaluation of how alternative decision policies behave throughout the drilling process, rather than evaluating performance solely from the final well trajectory.
\end{abstract}

\begin{graphicalabstract}
\includegraphics[width=1.1\linewidth]{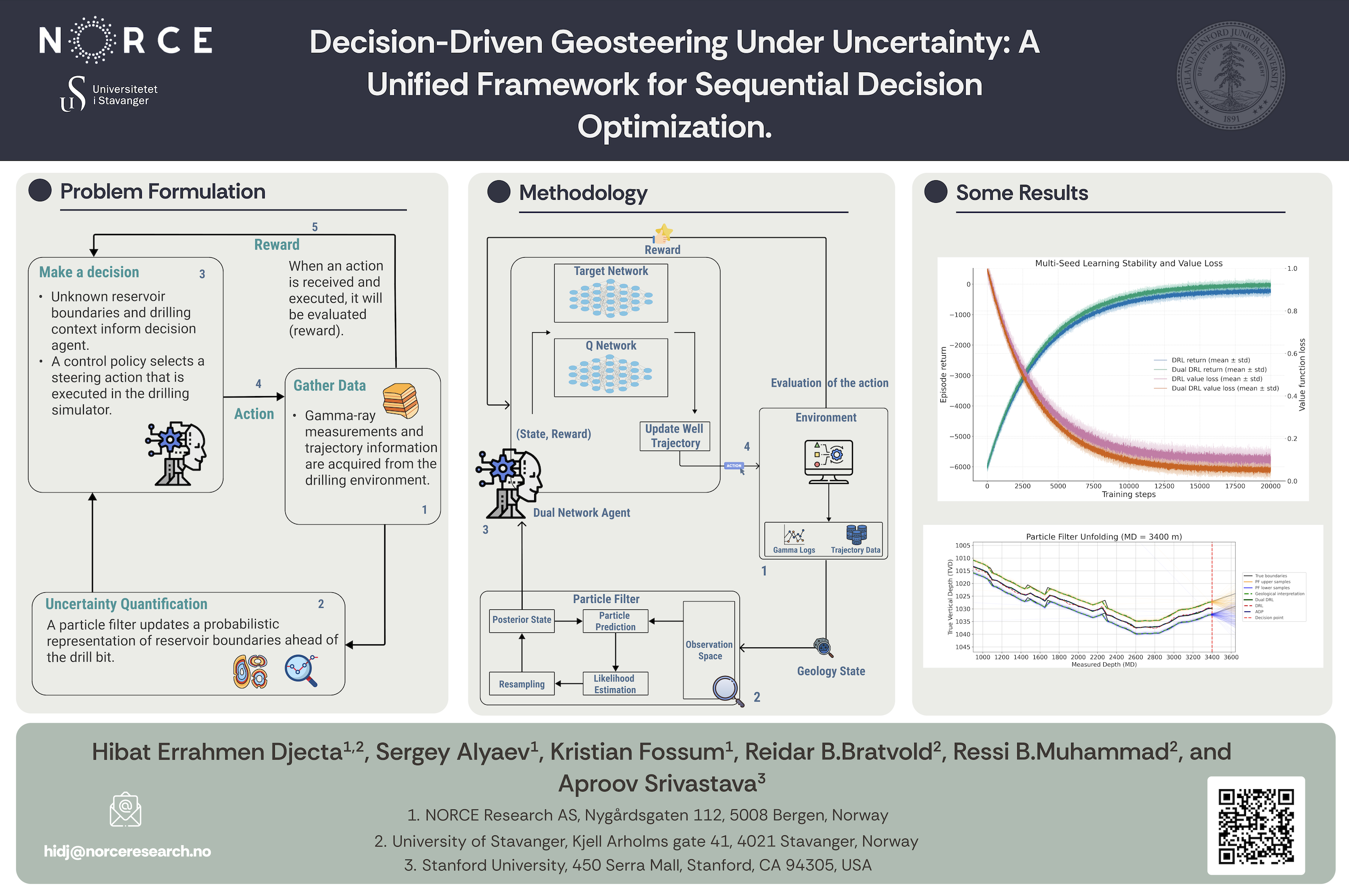}

\end{graphicalabstract}



\end{frontmatter}



\section{Introduction}
\label{sec1}

Sequential decision-making under uncertainty arises in many engineering applications, often illustrated by problems such as autonomous driving. In geosteering, this challenge is even more pronounced, as well trajectories must be adjusted in real time based on indirect observations of a largely unobservable subsurface. The objective is to maximize economic value by maintaining optimal reservoir contact while minimizing drilling costs and avoiding geological or operational hazards. These decisions rely on noisy measurements and incomplete geological information, making the problem inherently uncertain. Conventional geosteering workflows depend heavily on manual interpretation of real-time data, a process that is time-consuming, difficult to scale, and sensitive to human judgment. As a result, recent advances in artificial intelligence and computational science have driven growing interest in data-driven geosteering approaches that aim to formalize and improve sequential decision-making under geological uncertainty.

Kullawan et al.~\cite{10.2118/167433-PA} introduced a decision-analytic framework for geosteering that explicitly balances multiple objectives, including reservoir contact, cost control, and operational constraints. Their approach combined greedy optimization with a Bayesian representation of geological boundary uncertainty, thereby linking subsurface uncertainty to the sequential decision-making process. A subsequent study~\cite{KULLAWAN201890} extended this framework by applying Discretized Stochastic Dynamic Programming (DSDP) to fixed-thickness reservoirs, reporting improvements in well value of up to 31\% relative to the original method. Building on these approaches, Alyaev et al.~\cite{Alyaev_2019} proposed a Decision Support System (DSS) that integrated Ensemble Kalman Filtering (EnKF) \cite{Evensen2003} with simplified dynamic programming, enabling reproducible and effective decisions under uncertainty. More recently, Alyaev et al.~\cite{alyaev2024distinguish} further extended this DSS by incorporating geological models based on Generative Adversarial Networks (GANs), allowing decision-making in more complex geological settings.

In parallel with these developments, Reinforcement Learning (RL) \cite{Sutton1998} has emerged as a promising paradigm for sequential geosteering decisions. Muhammad et al.~\cite{optimal} introduced Deep Q-Networks (DQN) \cite{DBLP:journals/corr/MnihKSGAWR13} for geosteering and demonstrated that they outperform earlier model-based strategies, including greedy optimization and DSDP, on benchmark cases from~\cite{10.2118/167433-PA,KULLAWAN201890}, while avoiding many of their implementation limitations. 

Particle filtering and, more broadly, Sequential Monte Carlo (SMC) methods have also been used directly as a Bayesian geosteering engine to track stratigraphic boundaries under uncertainty. In ~\cite{veettil2020bayesian} geosteering was formulated  as a Bayesian inference problem and applied SMC to sequentially update probabilistic boundary interpretations as new measurements arrive. This line of work is particularly relevant to our setting because it operationalizes uncertainty through an explicit posterior over boundary configurations, providing a principled mechanism for real-time uncertainty propagation and conditioning. Building on this probabilistic interpretation capability, subsequent work integrated PF outputs into sequential decision policies, including the geosteering robot that couples multi-hypothesis PF interpretation with AI-based look-ahead decision-making~\cite{high}. This work confirmed the advantages of RL in more general and uncertain scenarios. Most recently, the authors adapted the PF+RL approach to a realistic Geosteering World Cup (GWC) environment, resulting in the so-called Pluralistic robot~\cite{geo}.

In post-GWC-2021 synthetic evaluations with noiseless measurements, the robot demonstrated performance comparable to, and in some cases exceeding, that of human experts, even though the PF was designed to operate under measurement noise. These results highlight the potential of reinforcement learning-based geosteering when probabilistic state estimation is reliable.

Nevertheless, important limitations remain. Existing geosteering approaches, including those based on reinforcement learning, have largely been validated in controlled synthetic settings that may not fully reflect the complexity and variability encountered in real drilling operations. While the Pluralistic robot successfully combined reinforcement learning with PF–based uncertainty modeling, it did not systematically investigate more stable Dual DRL architectures, nor did it assess performance under a broader range of real-time operational constraints.

This paper builds on the descriptions and experiments presented in our previous work \cite{10.1007/978-3-031-97554-7_14}. While the earlier study focused on demonstrating the feasibility and simulator-level performance of the dual-network architecture \cite{dual}, the present work introduces a unified uncertainty-quantification and decision framework for stratigraphic geosteering. The framework combines 
a data-driven geological generator based on a kernel-density estimator (KDE), a particle filter for uncertainty quantification and probabilistic stratigraphic forecasting and multiple robust sequential decision optimization options.
These options range from white-box approximate dynamic programming, to black-box dual reinforcement learning. 
Our implementation includes a simpler DQN method as a testing baseline.
The target is to improve explainability with the Approximate Dynamic Programming and increase robustness with Dual DRL. 
All three resulting policies are systematically analyzed for value gain, stability, and decision dynamics under geological uncertainty; we do so by refining the evaluation methodology, introducing stability-oriented performance metrics, and conducting a comparative analysis across multiple decision-making paradigms operating under identical PF-based uncertainty representations. This allows us to move beyond aggregate reward-based evaluation and instead characterize how different decision-making schemes respond to evolving uncertainty during sequential steering.

In addition, the framework is integrated with an API, enabling validation within a geosteering industrial simulator under realistic measurement noise models and operational constraints. This integration supports controlled yet high-fidelity testing across different geological scenarios, bridging the gap between synthetic experimentation and deployment-oriented evaluation.

The paper is organized as follows.
Section~\ref{sec:problem_formulation} formulates geosteering as a sequential decision-making problem under uncertainty. Section~\ref{sec:methodology} presents the PF framework and the decision-making methods considered, including ADP, standard Deep Reinforcement Learning (DRL), and the proposed Dual DRL approach. Experimental results and policy behavior analyses are reported in Section~\ref{sec:results}. Section~\ref{sec:discussion} discusses the implications of these results, and Section~\ref{sec:conclusion} concludes the paper.

\section{Problem Formulation}
\label{sec:problem_formulation}
Geosteering is an online decision-making problem in which a drilling trajectory must be continuously adjusted while advancing through a subsurface environment that is poorly known and only partially observable. Geological structures ahead of the drill bit are therefore unknown and can only be inferred indirectly from noisy measurements acquired during drilling. As a result, trajectory decisions must be made sequentially under uncertainty, balancing short-term placement quality against long-term objectives, while respecting strict operational constraints. This combination of partial observability, stochastic system evolution, and constrained control places geosteering within the class of sequential decision problems under uncertainty.

To formally capture these characteristics, the geosteering task is modeled as a Partially Observable Markov Decision Process (POMDP). In this formulation, the subsurface has a single geological configuration, but it cannot be directly observed while drilling. Instead, the decision-maker relies on indirect measurements obtained during drilling to update its knowledge of the subsurface and to select trajectory adjustments. The POMDP framework provides a principled mathematical structure for representing geological uncertainty, sequential information acquisition, and constrained control, and serves as the foundation for the decision-making problem addressed in this work.
It is defined by the tuple
\begin{equation}
\mathcal{M} = \langle \mathcal{S}, \mathcal{A}, \mathcal{O}, P, Z, R, \gamma \rangle,
\label{eq:pomdp_tuple}
\end{equation}
where each component is detailed below.

\subsection{Subsurface State}

Let $s_t \in \mathcal{S}$ denote the true subsurface state at drilling step $t$.
The state summarizes the configuration of the geological target interval relative to the wellbore, including quantities such as the vertical position, and local orientation of geological boundaries.

The subsurface state is assumed to evolve according to a Markovian stochastic process,
\begin{equation}
s_t \sim P(s_t \mid s_{t-1}, a_{t-1}),
\label{eq:state_transition}
\end{equation}
where the transition kernel $P(\cdot)$ (as defined in Eq.~\eqref{eq:pomdp_tuple}) captures geological variability, structural complexity, and modeling uncertainty.
The dependence on the previous action reflects the fact that trajectory decisions influence the spatial location at which new geological information is encountered. Although the subsurface is fixed, $P(\cdot)$ encodes epistemic uncertainty about
the unknown boundary geometry encountered next along the drilled path, rather
than physical stochastic evolution of geology.

Importantly, the true state $s_t$ is not directly observable during drilling, as geological structures ahead of the bit cannot be measured explicitly.

\subsection{Observations}

At each drilling step $t$, measurements acquired around the drill bit generate an observation
\begin{equation}
o_t \in \mathcal{O}.
\end{equation}
Observations are probabilistically related to the true subsurface state through the observation model
\begin{equation}
o_t \sim Z(o_t \mid s_t).
\label{eq:observation_model}
\end{equation}

The observation set may include logging responses, directional measurements, and trajectory information.
Due to measurement noise, limited sensing range, and the absence of direct look-ahead measurements, observations provide only partial and indirect information about the geological configuration ahead of the drill bit.

Consequently, observations are subject to measurement error and limited sensing
range, and thus cannot uniquely determine the subsurface state; they must be interpreted probabilistically.

\subsection{Belief State}

Given the partial observability of the subsurface, decision-making cannot rely on the true state $s_t$.
Instead, it is based on a belief state $b_t$, defined as a probability distribution over possible subsurface states:
\begin{equation}
b_t(s) = p(s_t = s \mid o_{1:t}, a_{1:t-1}).
\label{eq:belief_def}
\end{equation}

The belief state represents the decision-maker’s current knowledge of the subsurface, conditioned on the complete history of observations and actions.
It evolves recursively through Bayesian filtering using the transition and observation models defined in Eqs.~\eqref{eq:state_transition}--\eqref{eq:observation_model},
\begin{equation}
b_t(s') \propto Z(o_t \mid s')
\int_{\mathcal{S}} P(s' \mid s, a_{t-1}) \, b_{t-1}(s)\, ds,
\label{eq:belief_update}
\end{equation}
where the proportionality constant ensures normalization.

The belief $b_t$ constitutes a sufficient information state for optimal control in partially observable environments, allowing the original POMDP to be reformulated as a fully observable decision problem in belief space.

\subsection{Actions and Operational Constraints}

At each drilling step, a steering action is selected, representing a directional adjustment of the well trajectory, such as changes in inclination or azimuth.

\begin{equation}
a_t \in \mathcal{A}
\end{equation}

where $\mathcal{A}$ is the action space introduced in the POMDP tuple in Eq.~\eqref{eq:pomdp_tuple}.

Actions are constrained by physical and operational limitations.
In particular, the dogleg severity (DLS) between successive actions must satisfy
\begin{equation}
\mathrm{DLS}(a_t, a_{t-1}) \le \mathrm{DLS}_{\max},
\end{equation}
where $\mathrm{DLS}_{\max}$ denotes the maximum allowable dogleg severity imposed by drilling equipment and safety requirements.
Here, DLS quantifies the magnitude of the change in wellbore direction between two consecutive trajectory segments per unit drilled length (measured depth). It is commonly reported in degrees per m or degrees per ft.

This constraint restricts the admissible action set and introduces coupling between consecutive decisions, further emphasizing the sequential nature of the problem.

\subsection{Reward and Optimization Objective}

At each drilling step, a scalar reward is assigned to quantify geological performance.

\begin{equation}
r_t = R(s_t, a_t)
\label{eq:reward_general}
\end{equation}

where $R$ is the reward component of the POMDP in Eq.~\eqref{eq:pomdp_tuple}.

The reward function \cite{NIPS2000_e0ab531e} reflects the quality of well placement relative to the target interval and is defined as
\begin{equation}
r_t =
w_c \cdot \mathrm{contact}_t
-
w_o \cdot \mathrm{out\_of\_zone}_t,
\end{equation}
where $\mathrm{contact}_t$ measures the extent of reservoir contact and $\mathrm{out\_of\_zone}_t$ penalizes deviations outside the target zone.
The weights $w_c$ and $w_o$ control the relative importance of these objectives.

In our implementation, the reward is formulated on a penalty (cost) scale with an upper bound close to zero. As a result, cumulative returns are typically negative; better performance corresponds to values closer to zero (i.e., less negative). Specifically, $\mathrm{contact}_t$ is defined as a non-positive shaping term that reflects in-zone placement quality by penalizing distance from the target midpoint, so it approaches its maximum (closest to zero) when the trajectory remains near the interval center. Conversely, $\mathrm{out\_of\_zone}_t$ captures target-interval violation (zero inside the interval and increasing outside), and the term $-w_o\cdot \mathrm{out\_of\_zone}_t$ introduces an additional negative penalty to strongly discourage out-of-zone drilling. Consequently, negative rewards should be interpreted as deviations from the desired drilling behavior rather than as economic loss.

The goal is to determine a policy $\pi$ that maps belief states to actions,
\begin{equation}
a_t = \pi(b_t),
\label{eq:policy_on_belief}
\end{equation}
and to optimize this policy with respect to the objective in Eq.~\eqref{eq:objective}, where $\gamma$ is the discount factor introduced in Eq.~\eqref{eq:pomdp_tuple}.

\begin{equation}
\max_{\pi} \;
\mathbb{E}_{\pi}
\left[
\sum_{t=0}^{T}
\gamma^t r_t
\right],
\label{eq:objective}
\end{equation}
where $\gamma \in (0,1)$ is a discount factor and $T$ denotes the finite decision horizon.

\section{Methodology}
\label{sec:methodology}
Having formalized geosteering as a sequential decision-making problem under partial observability, we now describe the methodological framework.

This section describes the proposed geosteering decision-making framework built on probabilistic subsurface interpretation and learning-based control.
The methodology operates on belief-informed representations of geological uncertainty and integrates value-based decision-making mechanisms for sequential trajectory optimization.
Subsurface inference and control are treated as separate but tightly coupled components, enabling uncertainty-aware decisions while preserving physical and operational realism.

\subsection{System Overview}

At each drilling step, the framework executes a closed-loop sequence:
\begin{enumerate}
    \item Log measurements and trajectory information are acquired from the drilling environment.
    \item A particle filter updates a probabilistic representation of reservoir boundaries ahead of the drill bit.
    \item A decision state is constructed from the inferred boundary uncertainty and drilling context.
    \item A control policy selects a steering action that is executed in the drilling simulator.
\end{enumerate}
This loop continues until the target section is fully drilled or the decision horizon is reached.

\subsection{Particle Filter for Boundary Inference}
\label{subsec:pf_method}

Geological uncertainty is represented by a weighted particle ensemble, while the PF \cite{6714080} conditions this ensemble on incoming measurements (gamma-ray (GR) measurements) to refine the uncertainty over reservoir boundaries. The PF maintains a discrete set of particles, each corresponding to a plausible boundary configuration characterized by a vertical offset and local inclination.

Each particle state at drilling step $t$ is defined as
\begin{equation}
\xi_t^i = \bigl( \text{offset}_t^i,\; \text{angle}_t^i \bigr),
\end{equation}
where $\text{offset}_t^i$ denotes the vertical displacement of the target boundary relative to a reference depth, and $\text{angle}_t^i$ represents the local dip of the boundary. This low-dimensional parameterization captures the dominant geometric uncertainty relevant for real-time geosteering decisions.
To avoid confusion with the true subsurface state $s_t$ in Section~\ref{sec:problem_formulation}, we denote the PF particle state by $\xi_t^i$.

Particle states evolve according to a stochastic transition model,
\begin{equation}
\xi_t^i = f\bigl(\xi_{t-1}^i\bigr) + \varepsilon_t^i,
\label{eq:pf_transition}
\end{equation}

where $\varepsilon_t^i$ captures geological variability between successive drilling steps. In particular, angle increments are sampled from a proposal distribution defined by a KDE \cite{chen2017tutorialkerneldensityestimation} constructed from reference angle statistics. 
The KDE is trained offline by extracting a large sample of empirical angle-increment values from a reference set of stratigraphic interpretations (computed as local changes in boundary dip between successive discretization points/segments) and fitting a kernel density estimator to this one-dimensional sample of increments. 
This KDE-based transition model enables the PF to generate realistic boundary evolutions while preserving variability observed in historical or synthetic geological data.

Upon receiving a new GR measurement $o_t$, particle weights are updated according to
\begin{equation}
w_t^i \propto w_{t-1}^i \, p\bigl(o_t \mid \xi_t^i\bigr),
\label{eq:pf_weight_update}
\end{equation}
where $p\bigl(o_t \mid \xi_t^i\bigr)$ evaluates the likelihood of the observed GR measurement given the boundary configuration represented by particle $i$.
In practice, this likelihood is computed by comparing the measured GR sequence around the bit to a reference (offset-well) GR log that is shifted and locally aligned according to the boundary position implied by $\xi_t^i$ (i.e., the particle defines a mapping between measured depth and the corresponding reference-log depth), so particles that yield better log agreement receive higher weight.

Following weight normalization, resampling is performed when particle degeneracy is detected, ensuring that the particle set remains representative of the posterior distribution. The resulting weighted ensemble provides a real-time probabilistic estimate of reservoir boundary position and associated uncertainty, which is subsequently used by the decision-making module to evaluate and select steering actions under uncertainty.

\subsection{Decision State Representation}

Rather than operating directly on the full particle ensemble, a compact decision state is constructed to support learning-based control.
At each step, the decision state is defined as
\begin{equation}
x_t = \Phi\!\left( \{ \xi_t^i, w_t^i \}_{i=1}^{N_{\text{eff}}},\; o_t,\; a_{t-1} \right),
\label{eq:decision_state}
\end{equation}
where $\{ \xi_t^i, w_t^i \}_{i=1}^{N_{\text{eff}}}$ denotes a subset of the most informative particles (typically the top ones by weight; $N_{\text{eff}}$
denotes the number of informative particles),
$o_t$ is the current GR measurement, and $a_{t-1}$ is the previously applied steering action.

The feature mapping $\Phi(\cdot)$ extracts physically meaningful quantities from the PF posterior and recent trajectory history to preserve interpretability. Concretely, it includes (i) a local trajectory-context term given by the current inclination (summarizing the recent depth trend), (ii) the posterior weights of the five highest-probability particle hypotheses, and (iii) boundary-relative geometry descriptors evaluated at $26$ checkpoints spanning the most recent decision interval. At each checkpoint and for each retained hypothesis, $\Phi(\cdot)$ encodes two normalized signed distances between the observed well depth and the hypothesis-implied top and base boundaries depth (referenced to the interpreted horizon and scaled by a thickness parameter), capturing whether the trajectory lies above or below the boundary and by how much. This construction results in a fixed-length feature vector of $1 + 5\bigl(1 + 2\times 26\bigr)=266$ components, comprising one inclination feature, five posterior-weight features (one per retained hypothesis), and $5\times 2\times 26$ boundary-relative distance features evaluated across the checkpoints.

\subsection{Action Space Definition}

Steering actions are discretized according to drilling phases. 
During the landing phase, the agent selects inclination adjustment commands
\(
\Delta \theta
\)
from a discrete set spanning the range \(-10^\circ\) to \(+10^\circ\) with
increments of \(0.5^\circ\). These actions define how the wellbore is steered
toward the target reservoir interval during the build-up section. During the horizontal drilling phase, the agent selects from a set of discrete
vertical target-line adjustments that shift the planned well trajectory upward
or downward relative to the interpreted target interval. These actions are defined as discrete vertical offsets applied to the forward target-line control points and, in our implementation, take values in the range $-6$ to $+6$ (with a step size of $0.5$ in depth units). They are designed to keep the wellbore within the productive interval while promoting smooth steering and avoiding unnecessary directional changes. Low-level drilling mechanics, including inclination execution, trajectory
smoothing, and wellbore curvature, are handled by the drilling environment.
In particular, rapid changes in direction are restricted by a
DLS constraint, which limits the maximum allowable curvature of the wellbore
to protect the drilling assembly. Actions that would violate this constraint are
filtered by the environment and are therefore not executed.
Because the geological impact of a steering decision may only become observable
several drilling steps later, the agent must operate under delayed feedback,
requiring robust sequential decision-making rather than myopic control.

\subsection{Learning-Based Value Estimation}

Decision-making is guided by a state–action value function
\begin{equation}
Q(x_t, a_t),
\end{equation}
which estimates the expected cumulative discounted reward following action $a_t$ in state $x_t$.
Learning is performed using Temporal-Difference updates based on observed transitions $(x_t, a_t, r_t, x_{t+1})$.

Experience replay is employed to improve sample efficiency and stabilize training by reducing temporal correlations between updates.

\subsection{Approximate Dynamic Programming (ADP)}

To provide a structured, white-box non-learning option, we implement an ADP approach.

ADP represents a class of methods that approximate the optimal value function using a fixed parametric or heuristic model, without relying on trial-and-error learning through interaction.

The optimal value function
\begin{equation}
V^\ast(x)
=
\max_{\pi}
\mathbb{E}_{\pi}
\left[
\sum_{t=0}^{T}
\gamma^t r_t
\;\middle|\;
x_0 = x
\right]
\end{equation}
is approximated by a surrogate function
\begin{equation}
\hat{V}(x) \approx V^\ast(x),
\end{equation}
where $x$ denotes the belief-informed decision state.
The approximation $\hat{V}(x)$ is constructed using domain-informed features extracted from the belief state, such as estimated boundary position, uncertainty bounds, and local trajectory context.

Given the approximate value function, actions are selected using a one-step look-ahead policy:
\begin{equation}
a_t
=
\arg\max_{a \in \mathcal{A}_t}
\left[
R(x_t, a)
+
\gamma \, \hat{V}(x_{t+1})
\right],
\end{equation}
where $x_{t+1}$ denotes the predicted next decision state resulting from applying action $a$ at state $x_t$.
This formulation explicitly accounts for immediate geological reward and a discounted estimate of future value, while avoiding full dynamic programming over long horizons.

Because geological uncertainty plays a central role in geosteering, the ADP incorporates an optimism mechanism to encourage informative exploration.
Specifically, the value estimate is augmented as
\begin{equation}
\hat{V}_{\text{opt}}(x)
=
\hat{V}(x)
+
\beta \, \sigma(x),
\end{equation}
where $\sigma(x)$ denotes a scalar uncertainty measure derived from the belief state, and $\beta > 0$ controls the degree of optimism.

This optimistic bias favors actions that either improve expected geological performance or reduce subsurface uncertainty, reflecting classical decision-analytic principles of value of information.
Importantly, this mechanism does not rely on stochastic exploration or learning, but instead exploits belief-derived uncertainty directly in the control policy.

The ADP method provides a computationally efficient, interpretable option that leverages the same belief representation and action constraints as the learning-based approaches.
However, because the value approximation is fixed and does not improve through interaction, ADP lacks the ability to adapt to long-term reward structure or delayed consequences.
This limitation makes ADP particularly suitable as a benchmark for assessing the benefits of learning-based value estimation under geological uncertainty.

\subsection{Dual Deep Reinforcement Learning (Dueling Deep Q-Network)}
\label{subsec:dueling_dqn}
To improve learning stability under partial observability, long decision horizons, and noisy belief updates, we employ a dueling value-based architecture \cite{dual}. The key idea is to represent the state--action value function as a combination of (i) a state-value estimator and (ii) an advantage estimator. This is particularly beneficial when many actions have similar long-term outcomes, since the network can learn how desirable a belief-informed state is even when action advantages are small.

Let $x_t$ denote the belief-informed decision state (Eq.~\eqref{eq:decision_state}), $r_t$ the scalar reward (Eq.~\eqref{eq:reward_general}), and $\gamma \in (0,1)$ the discount factor from the POMDP tuple in Eq.~\eqref{eq:pomdp_tuple}. The dueling network computes a scalar value $V_{\theta}(x_t)$ and an advantage vector
$A_{\theta}(x_t,a)$ over actions, which are aggregated to form $Q$-values via the mean-normalized combination
\begin{equation}
Q_{\theta}(x_t,a)
=
V_{\theta}(x_t)
+
\left(
A_{\theta}(x_t,a)
-
\frac{1}{|\mathcal{A}|}\sum_{a'} A_{\theta}(x_t,a')
\right).
\label{eq:dueling_aggregation}
\end{equation}
This aggregation ensures identifiability between value and advantage components, and yields a stable estimate of $Q_{\theta}(x_t,a)$ for action selection.
At each decision step, the agent observes a transition $(x_t, a_t, r_t, x_{t+1})$. Training follows standard DQN updates with experience replay and a target network for stability. The temporal-difference target is computed using the target network $Q_{\bar{\theta}}$ as
\begin{equation}
y_t = r_t + \gamma \max_{a'} Q_{\bar{\theta}}(x_{t+1}, a'),
\label{eq:dqn_target}
\end{equation}
and the online parameters $\theta$ are updated by minimizing the squared
temporal-difference error
\begin{equation}
\mathcal{L}(\theta) = \bigl(y_t - Q_{\theta}(x_t, a_t)\bigr)^2.
\label{eq:td_loss}
\end{equation}
Experience replay improves sample efficiency and reduces temporal correlations between successive updates, which is important in geosteering where consecutive decisions are strongly coupled.

The target network parameters $\bar{\theta}$ are updated more slowly to provide a stable learning target. A soft update rule is used,
\begin{equation}
\bar{\theta} \leftarrow \tau \theta + (1 - \tau)\bar{\theta},
\label{eq:soft_update}
\end{equation}
where $\tau \ll 1$ controls the rate at which the target network tracks the online network. 

\subsection{Sequential Integration of PF and Decision Policies}

\begin{figure}[ht]
    \centering
    \fbox{\includegraphics[width=1\textwidth]{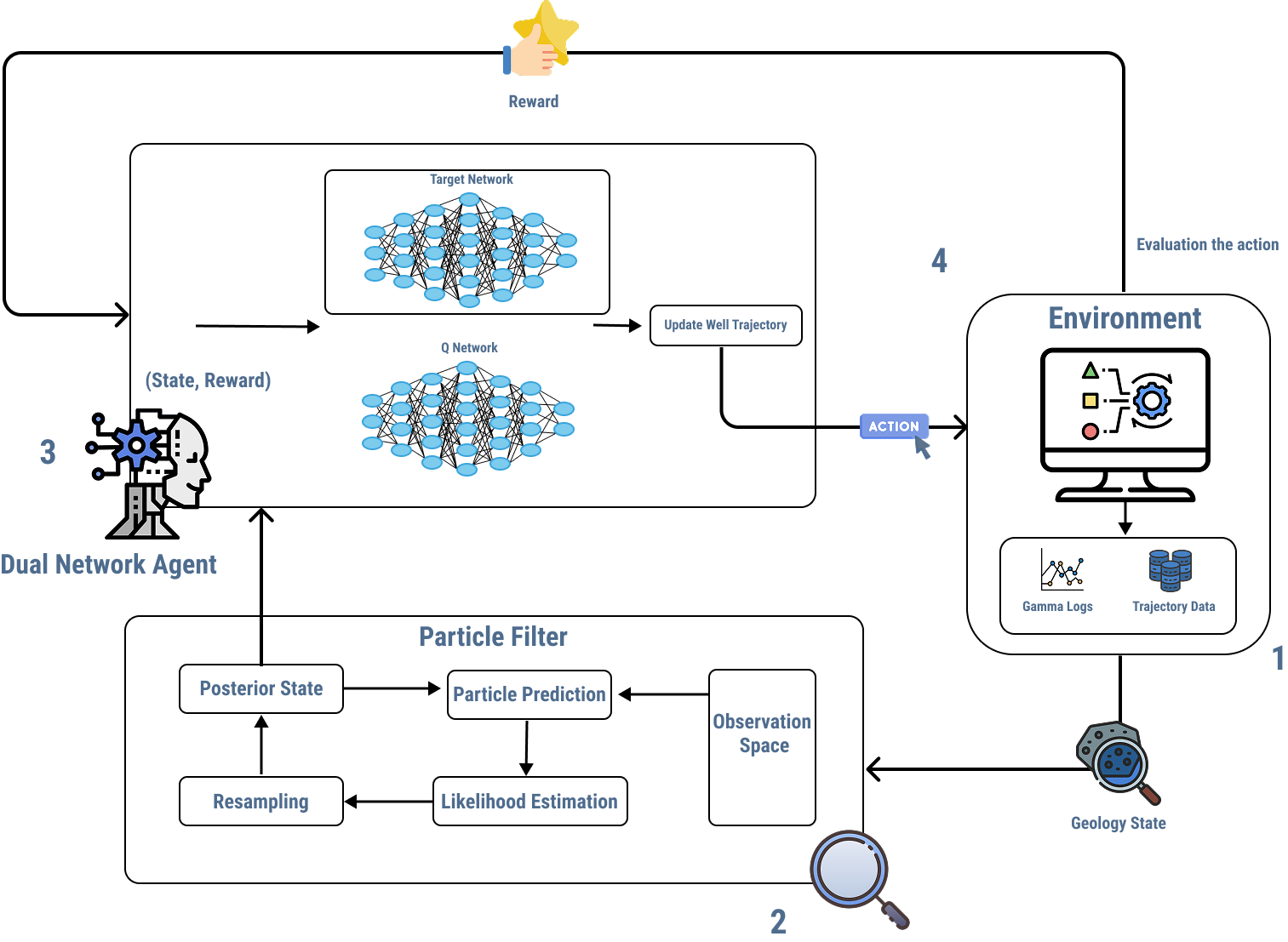}}
    \caption{High-level architecture illustrating the sequential data flow among the PF, Dual DRL, and the drilling environment. Uncertainties in geological parameters are handled by the PF, while the DRL agent selects optimal actions \cite{10.1007/978-3-031-97554-7_14}.}
    \label{fig:system_architecture}
\end{figure}

At each drilling step, the PF refines its posterior distribution of reservoir boundary configurations using incoming GR measurements.
The decision state constructed from this posterior is passed to the control policy, which selects a steering action executed in the drilling simulator.
This sequential coupling enables uncertainty-aware control while preserving real-time performance (see Figure~\ref{fig:system_architecture} for the Dual DRL case).

\subsection{Performance and Stability Metrics}

The proposed framework is evaluated using metrics that assess both geological placement accuracy and the stability of sequential steering decisions. This dual perspective is essential in geosteering, where aggressive short-term corrections may improve instantaneous placement while degrading overall trajectory smoothness and operational reliability.

To evaluate the smoothness and stability of steering decisions, we quantify
the rate of change of curvature in the vertical profile using the discrete jerk of the trajectory in true vertical depth (TVD).
Let $T$ denote the total number of discretized trajectory steps in the evaluated interval (i.e., the final drilling step index), and $z_t$ denote the TVD of the well at step $t$, and let $\Delta s$ be the
(step) increment along the trajectory (e.g., measured depth).
The discrete jerk $j_t$ is defined by the third-order finite difference
\begin{equation}
j_t
=
\frac{z_t - 3z_{t-1} + 3z_{t-2} - z_{t-3}}{\Delta s^3},
\qquad t = 4,\ldots,T.
\end{equation}
Overall control smoothness is summarized using the root mean square (RMS) jerk \cite{6ce60d164cfb4c5cb00bfc2a662a8e4d},
\begin{equation}
\mathrm{Jerk}_{\mathrm{RMS}}
=
\sqrt{
\frac{1}{T-3}
\sum_{t=4}^{T}
j_t^2
}.
\end{equation}
Lower RMS jerk indicates smoother trajectory adjustments, i.e., fewer abrupt
changes in steering that can translate into more stable directional control and
reduced tool loading in practice.

\section{Experimental setup and results}  \label{sec:results}

This section evaluates the proposed learning-based geosteering framework under geological uncertainty. All methods are assessed using identical geological realizations, reward definitions, operational constraints, and evaluation protocols, ensuring a controlled and fair comparison. Performance differences therefore arise exclusively from the underlying decision-making and learning mechanisms rather than from environmental or experimental variability.

All experiments were conducted on a high-performance computing (HPC) system equipped with a 13th Gen Intel® Core™ i7-13800H (20 threads), 32~GB system memory, and an NVIDIA GPU, and running Ubuntu 22.04. Geological uncertainty was represented through probabilistic sampling of subsurface realizations using a KDE–based model, enabling consistent exposure to stochastic geological conditions across methods. Unless otherwise stated, results are averaged over multiple random seeds, and all runs use the fixed hyperparameter set summarized in Table~\ref{tab:parameters}. This ensures consistency across experiments and allows observed performance differences to be attributed solely to the underlying learning algorithms rather than to variations in training configuration.

In addition, all experiments rely on GR measurements and trajectory data \cite{autosteering} as subsurface observations.

For testing, the StarSteer simulator was employed via Solo API \cite{SoloAPI} to communicate new placements and extract feedback from real subsurface interactions, allowing the robot to operate in a dynamic and realistic setting. 
Such a setup ensures an unbiased evaluation process for the developed method.

\subsection{Learning Dynamics of DRL-Based Methods}
\label{subsec:learning_dynamics}

We begin by analyzing the learning dynamics of the two learning-based approaches, DRL and Dual DRL. All results in this subsection are obtained under identical training horizons, reward definitions, and geological realizations, and are averaged over multiple random seeds.

We first investigate the impact of replay buffer capacity on learning stability. The replay buffer configurations considered in this study are summarized in Table~\ref{tab:replay_buffer_sizes}.

\begin{table}[t]
\centering
\caption{Replay buffer configurations evaluated in the learning dynamics experiments.}
\label{tab:replay_buffer_sizes}
\begin{tabular}{lcc}
\hline
Configuration & Buffer size (transitions) & Usage in experiments \\
\hline
Small buffer  & 20{,}000 & Sensitivity analysis only \\
Medium buffer & 50{,}000 & Default setting \\
Large buffer  & 70{,}000 & Sensitivity analysis only \\
\hline
\end{tabular}
\end{table}

\begin{figure}[t]
    \centering
    \includegraphics[width=\linewidth]{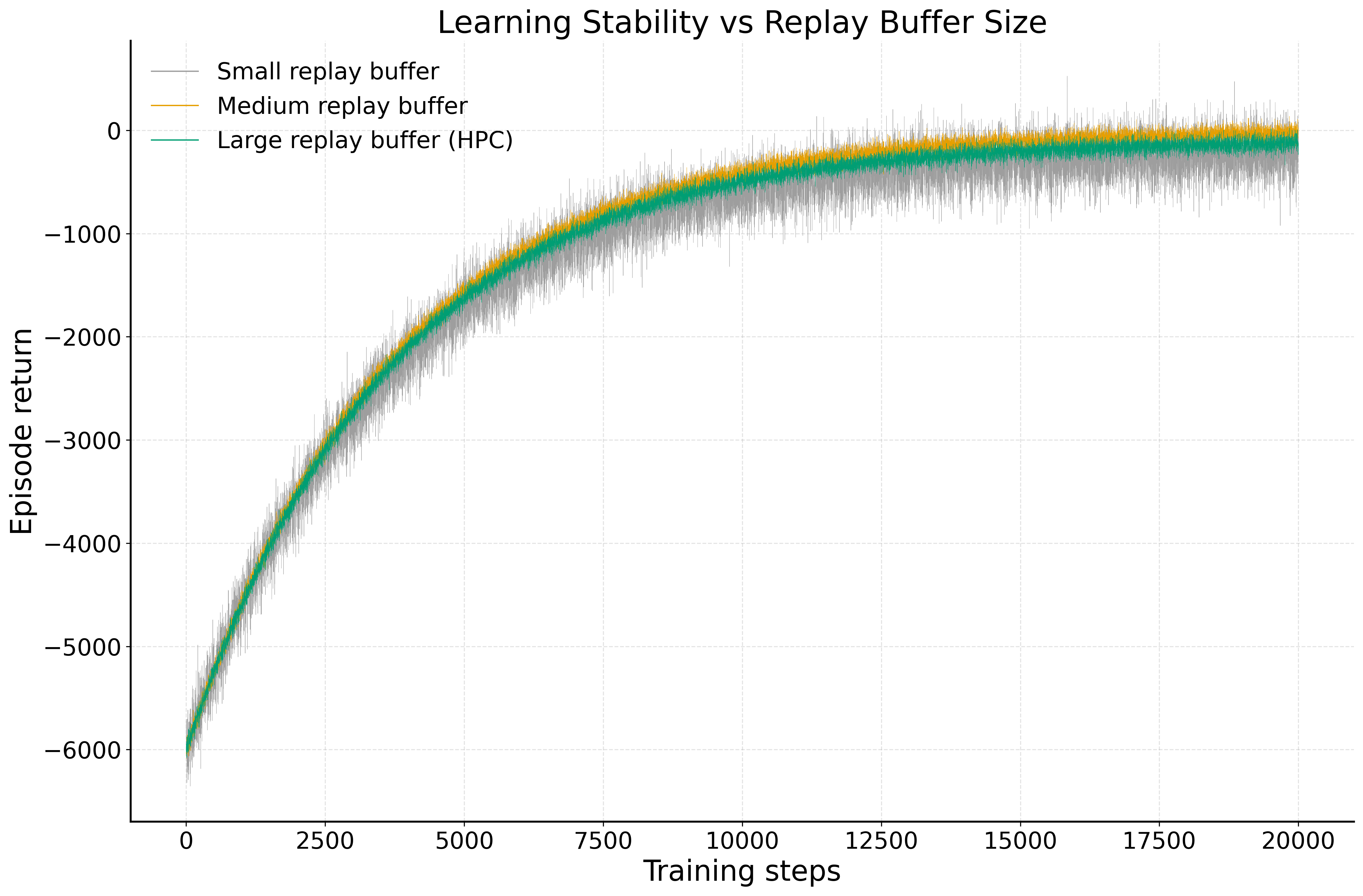}
    \caption{Learning dynamics under different replay buffer capacities (mean $\pm$ standard deviation across random seeds).
Episode return is reported in the original reward scale (higher is better). Smaller buffers yield higher variance during training, while larger buffers smooth learning by increasing experience diversity.}

    \label{fig:replay_buffer}
\end{figure}

Figure~\ref{fig:replay_buffer} compares learning dynamics obtained with different replay buffer capacities. The smallest replay buffer exhibits rapid early improvements but also displays increased variability throughout training, reflecting strong temporal correlations and limited diversity in stored experience. Such behavior is indicative of unstable value updates driven by repeated reuse of recent transitions.

Increasing the replay buffer size improves learning stability by broadening the distribution of stored experience and reducing variance in value estimation. The medium-sized buffer achieves the most favorable balance, combining stable convergence with consistently higher asymptotic performance. While the largest buffer yields the smoothest learning trajectory, its final performance does not improve further and exhibits slower adaptation due to the increasing influence of older, less relevant transitions.

The feasibility of evaluating larger replay buffers is enabled by the availability of high-performance computing (HPC) resources, which provide sufficient memory capacity and parallelized data handling to support extended experience storage. However, while HPC allows the use of larger buffers, the results indicate that increased capacity alone does not guarantee improved learning outcomes. Beyond a certain size, the benefits of additional experience diversity diminish, and learning efficiency may be reduced due to slower policy adaptation.

Based on this trade-off, a replay buffer size of 50{,}000 transitions is selected as the default configuration for all subsequent experiments in this subsection and throughout the remainder of the paper.

\begin{figure}[t]
    \centering
    \includegraphics[width=\linewidth]{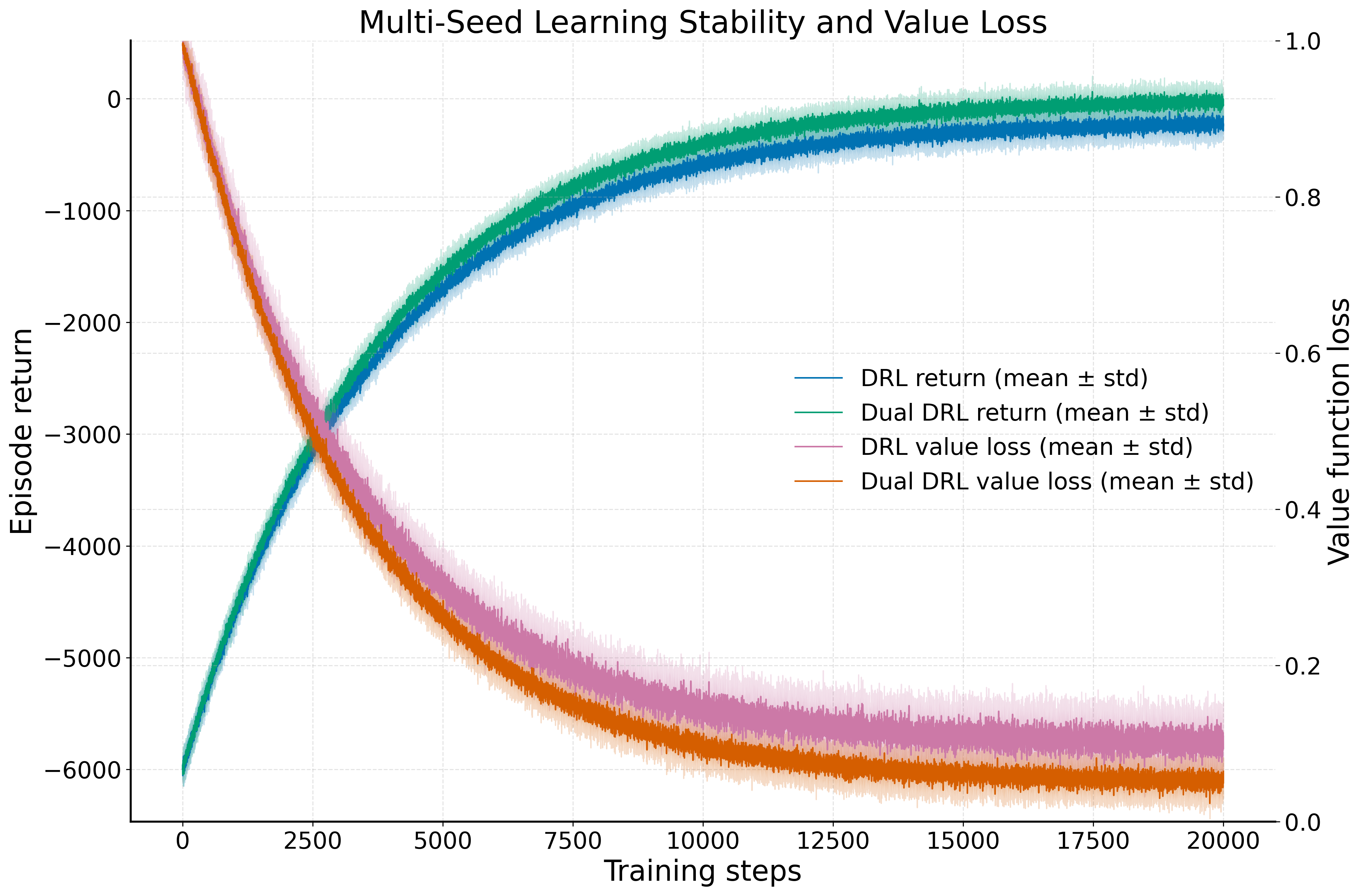}
\caption{Multi-seed learning stability for DRL and Dual DRL (mean $\pm$ standard deviation across random seeds).
Episode return (left axis; higher is better) and normalized value-function loss (right axis) are shown over training. Dual DRL exhibits reduced variance across seeds and smoother optimization dynamics.}

    \label{fig:multiseed_learning}
\end{figure}

Using this fixed replay buffer size, we next evaluate learning robustness across multiple random initializations. Figure~\ref{fig:multiseed_learning} reports the mean episode return and corresponding standard deviation across seeds for both DRL and Dual DRL, together with the normalized value-function loss shown on a secondary axis.

Both methods demonstrate consistent improvement in episode return as training progresses. However, Dual DRL exhibits reduced variance across seeds throughout training, indicating improved robustness to stochastic initialization and exploration noise. The accompanying loss curves further reveal smoother and more stable value-function optimization for Dual DRL, suggesting improved numerical stability during training.

\begin{figure}[t]
    \centering
    \begin{minipage}{0.48\linewidth}
        \centering
        \includegraphics[width=\linewidth]{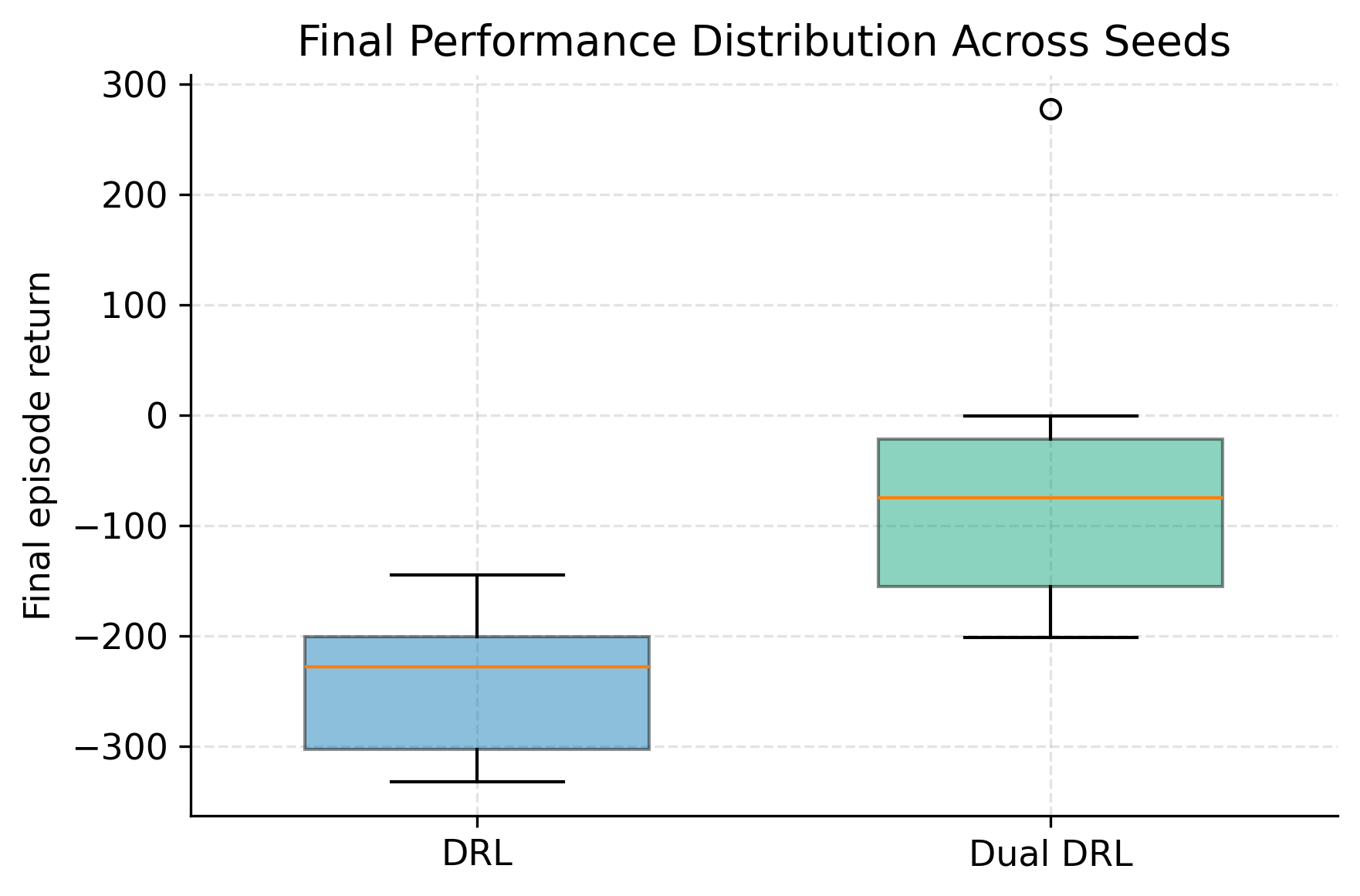}
        \caption{Distribution of final evaluation episode returns across independent training runs (random seeds; higher is better).
        Each sample corresponds to one trained model evaluated using the same protocol. Dual DRL achieves higher median performance with reduced dispersion compared to standard DRL.}
        
        \label{fig:final_distribution}
    \end{minipage}\hfill
    \begin{minipage}{0.48\linewidth}
        \centering
        \includegraphics[width=\linewidth]{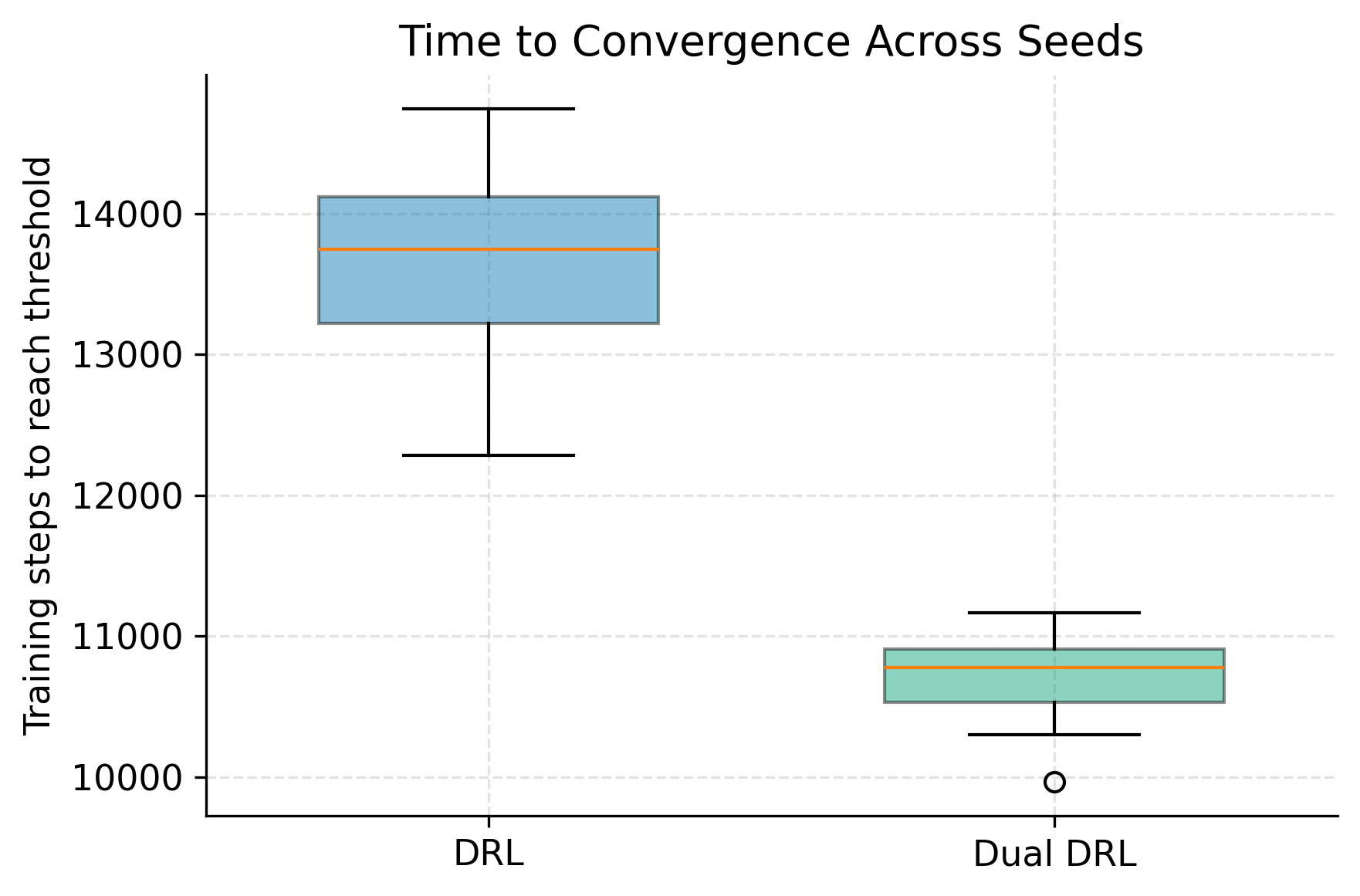}
        \caption{Training steps required to reach a fixed episode-return threshold across random seeds. Each point corresponds to one training run; lower is better. Dual DRL reaches the threshold earlier on average and with reduced variability, indicating more predictable learning behavior.}

        \label{fig:time_to_convergence}
    \end{minipage}
\end{figure}

The distribution of final episode returns across seeds is summarized in Figure~\ref{fig:final_distribution}. Dual DRL achieves a higher median performance with reduced dispersion, indicating that performance gains are consistent across independent training runs rather than driven by favorable initializations.

Complementarily, Figure~\ref{fig:time_to_convergence} reports the number of training steps required to reach a fixed performance threshold across seeds. Dual DRL converges earlier on average and with lower variability, demonstrating improved sample efficiency and more predictable learning behavior.

These properties provide a robust foundation for the comparative analysis with approximate dynamic programming methods presented in the following subsection.

\subsection{Decision Quality and Control Stability}
\label{subsec:decision_quality}

Beyond learning stability, an effective geosteering policy must yield reliable decision quality and smooth control behavior under geological uncertainty. In this subsection, we compare ADP, standard DRL, and Dual DRL in terms of final policy performance and control smoothness. Importantly, all reported statistics are computed over multiple independent runs (random seeds), ensuring that results reflect robustness rather than single-run behavior.

\begin{figure}[t]
    \centering
    \includegraphics[width=0.8\linewidth]{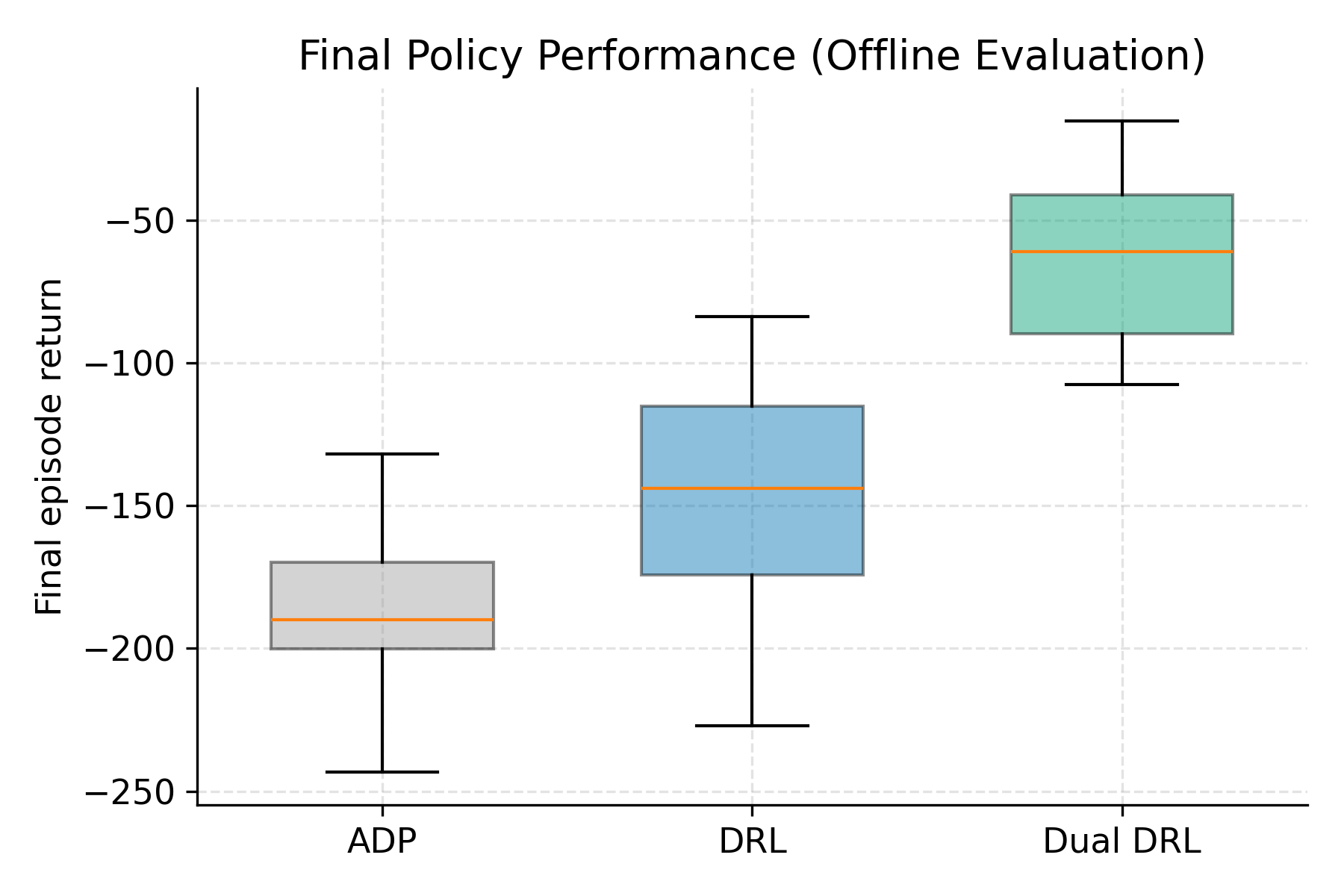}
    \caption{Final policy performance evaluated offline on a fixed evaluation set.
    Box plots show the distribution of evaluation episode returns across independent runs (random seeds) for ADP, DRL, and Dual DRL (higher is better). Medians and quartiles are computed over seeds, not over learning episodes.}

    \label{fig:final_performance}
\end{figure}

Figure~\ref{fig:final_performance} summarizes the final policy performance obtained by each method using box plots of episode returns. Each box represents the distribution of final episode returns across independent runs, with the median indicating the typical performance level. ADP exhibits the lowest median performance and a relatively narrow spread, reflecting its deterministic nature and limited capacity to improve beyond its fixed approximation structure. Standard DRL achieves higher median performance but shows increased variability across runs, indicating sensitivity to stochastic exploration and value estimation noise. Dual DRL attains the highest median return with reduced dispersion, demonstrating both improved performance and greater robustness across seeds.

To complement performance-based evaluation, we assess control smoothness using the RMS jerk. RMS jerk provides a physically meaningful proxy for steering aggressiveness and mechanical stress, with lower values indicating smoother and more stable control actions. Since ADP does not involve iterative learning, this metric is computed directly from its deployed policy and compared against the final trained policies of DRL and Dual DRL.

Table~\ref{tab:rms_jerk} reports the RMS jerk values averaged over independent runs. Dual DRL achieves a substantially lower RMS jerk than both ADP and standard DRL, indicating significantly smoother steering decisions. In our experiments, the DRL baseline exhibits slightly higher RMS jerk than ADP (1.38 vs.\ 1.32), reflecting more frequent local corrections. In contrast, Dual DRL produces smoother trajectories by stabilizing value updates and capturing longer-term decision structure.

\begin{table}[t]
    \centering
    \caption{Control smoothness measured by RMS jerk.
    Values are averaged over independent runs; lower is better.}
    \label{tab:rms_jerk}
    \begin{tabular}{lc}
        \hline
        Method & RMS jerk (3rd finite difference of TVD) \\
        \hline
        ADP & 1.32 \\
        DRL & 1.38 \\
        Dual DRL & \textbf{0.8} \\
        \hline
    \end{tabular}
\end{table}

\subsection{Policy Behavior Under Geological Uncertainty}
\label{subsec:policy_uncertainty}

This subsection analyzes how different decision-making policies behave under evolving geological uncertainty, as represented by the PF unfolding process. The analysis focuses on (i) the structure of uncertainty behind the drill bit after conditioning on observations, (ii) the propagation of uncertainty ahead of the bit in the absence of new measurements, and (iii) the resulting behavior of each policy under these conditions.

\begin{figure}[ht]
    \centering
    \includegraphics[width=0.85\linewidth]{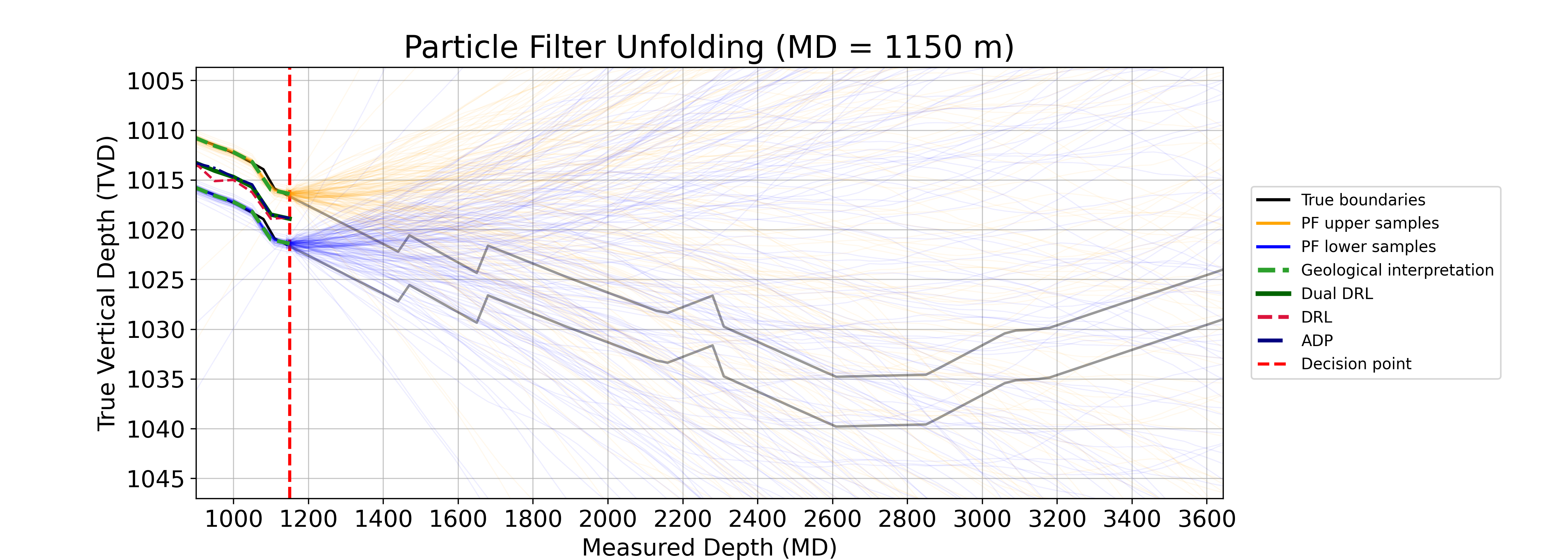}
    \includegraphics[width=0.85\linewidth]{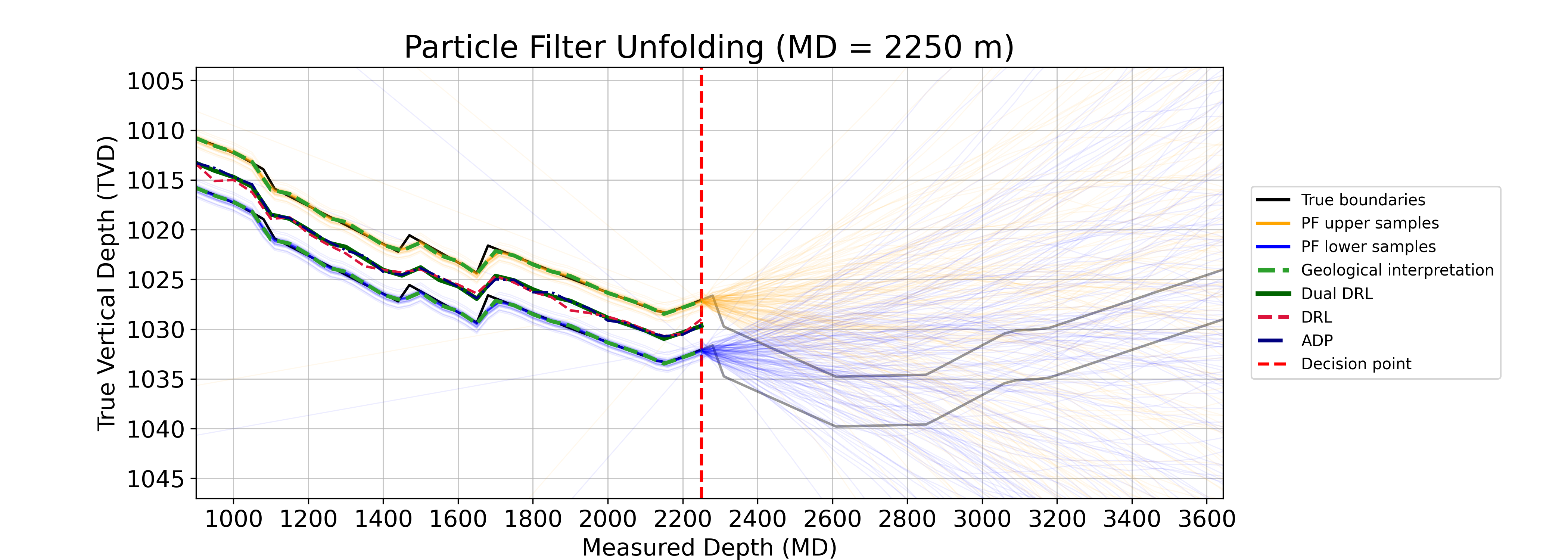}
    \includegraphics[width=0.85\linewidth]{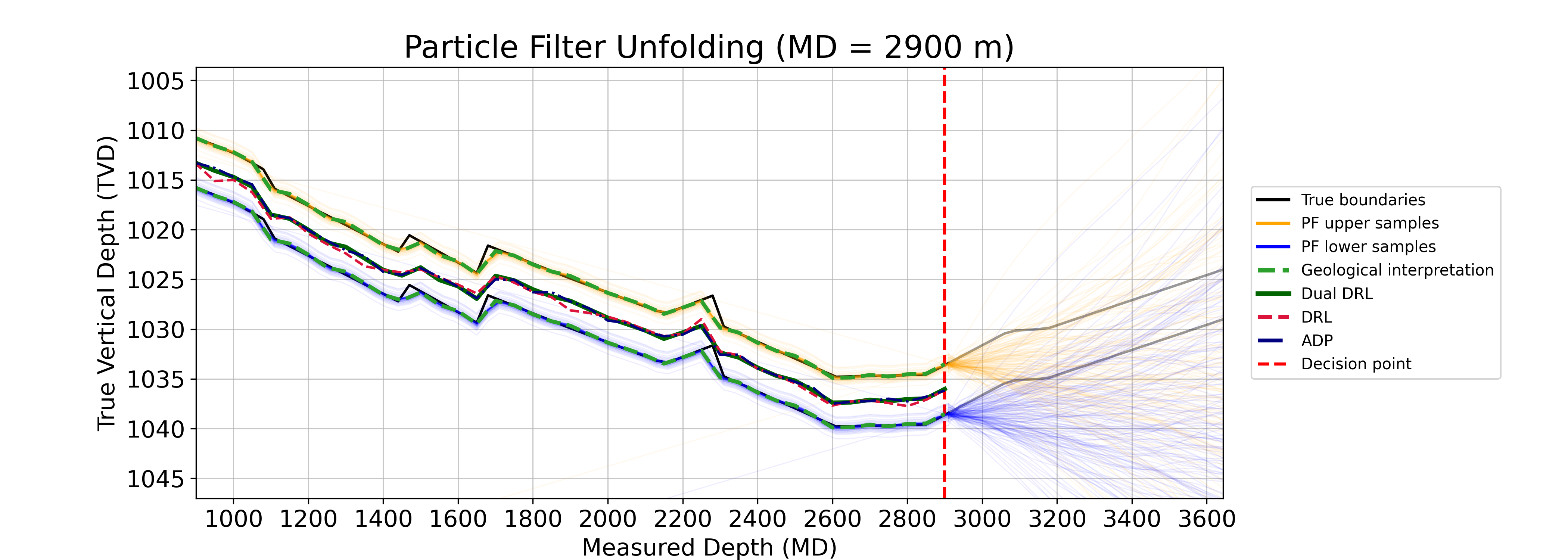}
    \includegraphics[width=0.85\linewidth]{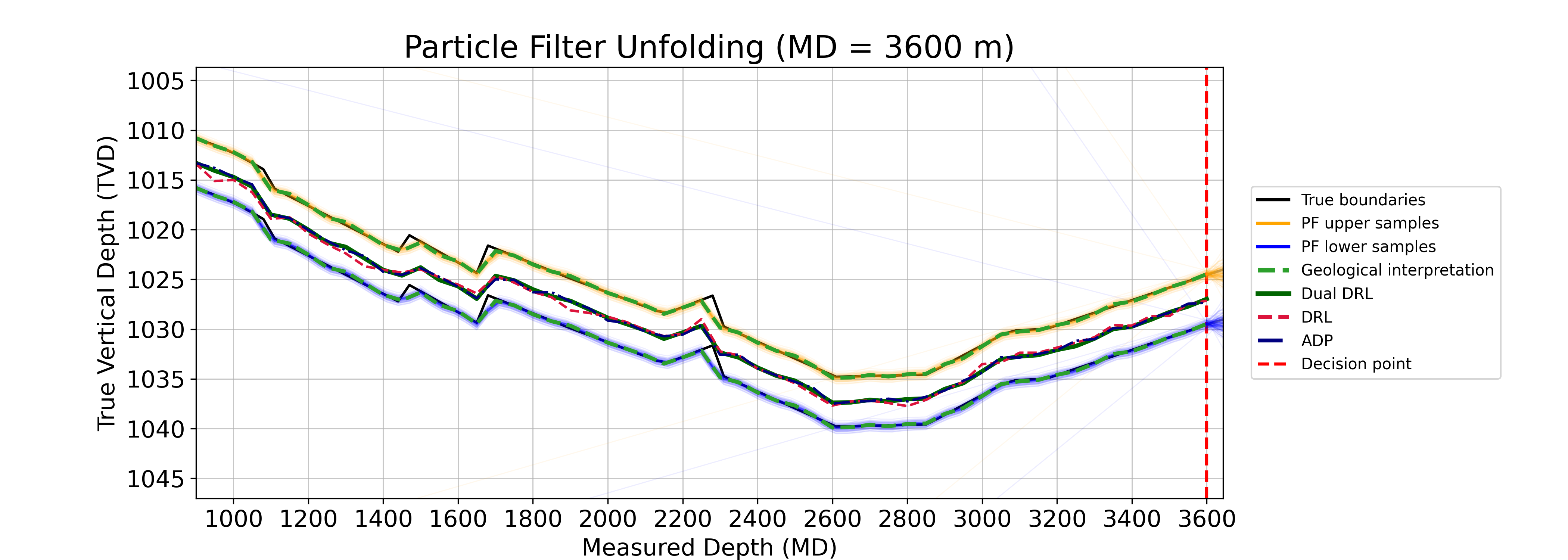}
    \caption{Particle-filter unfolding at four representative decision points along the well trajectory. The vertical dashed line marks the current decision point. Behind the bit, particles represent the posterior conditioned on accumulated observations; ahead of the bit, particles propagate without new measurements, forming a predictive uncertainty envelope. Overlaid trajectories show how ADP, DRL, and Dual DRL respond to the same evolving belief state (differences reflect policy behavior, not changes in the environment).}

    \label{fig:pf_unfolding_sequence}
\end{figure}

Figure~\ref{fig:pf_unfolding_sequence} illustrates four representative snapshots of the PF unfolding at different measured depths along the well trajectory. In each snapshot, the vertical dashed line denotes the current decision point. Behind this point, the PF is conditioned on the accumulated geological interpretation, while ahead of the bit the PF evolves freely according to the stochastic transition model.

Behind the bit, the posterior particle distribution exhibits a mixed structure. A subset of particles collapses tightly around the interpreted boundaries, reflecting strong conditioning from observed data. A second subset remains approximately parallel to the interpretation, capturing residual structural ambiguity, while a small number of particles diverge more strongly, representing low-probability geological alternatives. This heterogeneous posterior structure avoids unrealistic overconfidence while remaining consistent with the interpreted geology, as clearly visible in the left portions of each snapshot in Figure~\ref{fig:pf_unfolding_sequence}.

Ahead of the decision point, uncertainty unfolds gradually as particles originate from the same measured depth but from slightly different true vertical depths. These particles remain initially close to the interpretation while allowing slope variability, producing a fan-shaped predictive envelope that widens with distance from the bit. This behavior reflects the accumulation of geological uncertainty in the absence of new measurements, rather than numerical artifacts or policy-induced effects.

The drilling policies respond differently to this evolving uncertainty. The ADP policy exhibits oscillatory behavior around the interpreted centerline, particularly in regions of increased forward uncertainty. This behavior reflects a reactive correction mechanism that is sensitive to local deviations. The DRL policy produces moderately smooth trajectories, but shows more frequent local corrections than ADP, leading to slightly higher RMS jerk.

In contrast, the Dual DRL policy demonstrates the most stable behavior under geological uncertainty. As shown across all snapshots in Figure~\ref{fig:pf_unfolding_sequence} (zoomed-in in \ref{fig:pf_unfolding_sequence_400}), its trajectory remains closely aligned with the interpreted structure while avoiding excessive oscillations or abrupt corrections. This stability is especially evident in regions where forward uncertainty increases, indicating that the policy effectively balances responsiveness with robustness.

\section{Discussion}
\label{sec:discussion}

The results demonstrate that geosteering policy quality cannot be assessed solely from trajectory overlap or final performance metrics. Although ADP, DRL, and Dual DRL often produce visually similar paths, their behavior under geological uncertainty differs substantially. ADP exhibits reactive and oscillatory responses to local deviations, while standard DRL improves placement performance but remains sensitive to fluctuations in the inferred subsurface state. In contrast, Dual DRL consistently produces smoother and more stable decisions as uncertainty unfolds ahead of the drill bit. The PF unfolding highlights these differences by revealing how each policy responds to growing predictive uncertainty, emphasizing that stability and robustness are essential criteria for operationally viable geosteering.

\section{Conclusion}
\label{sec:conclusion}

This work examined learning-based geosteering under imperfect knowledge of subsurface boundaries within a unified uncertainty-quantification and decision framework. The framework couples a KDE-based geological generator with PF-based probabilistic stratigraphic forecasting, and uses the resulting belief-state representation to support multiple decision strategies. By coupling probabilistic boundary inference with learning-based control and complementary decision schemes, the analysis focused not only on final placement performance and value gain but also on policy behavior and stability as uncertainty evolves during drilling.

The results demonstrate that visually similar well trajectories can arise from fundamentally different decision policies, highlighting the limitations of evaluating geosteering performance based on final placement alone. Approximate Dynamic Programming exhibits reactive and oscillatory behavior, while the deep Q-learning baseline yields competitive placement but slightly higher control variability (RMS jerk) than ADP. In contrast, the proposed Dual DRL architecture consistently produces more stable and coherent decisions, particularly in regions of increasing predictive uncertainty.

PF unfolding proved valuable for exposing behavioral differences that are not captured by aggregate performance metrics alone, underscoring the importance of stability-aware evaluation for geosteering systems. This illustrates that differences in decision quality are primarily expressed through how policies interpret and act on evolving PF-based belief states, rather than through final trajectory geometry alone.

Future work will extend this framework to more complex geological settings and additional operational constraints, and will integrate sequence-level decision models into the same PF-based belief-state setup. Decision Transformer–based sequence-level geosteering has been studied separately \cite{chen2021decisiontransformerreinforcementlearning} in our recent work \cite{djecta2025geosteering}, but remains to be incorporated into the framework presented here.

\section*{Acknowledgments}

H.E. Djecta, S. Alyaev, K. Fossum, and R.B. Bratvold 
acknowledge the support from the project DISTINGUISH (Decision support using neural networks to predict geological uncertainties when geosteering), funded by Aker BP, Equinor, and the Research Council of Norway (RCN PETROMAKS2 project no. 344236).

R.B. Muhammad acknowledges the support from the Center for Research-based Innovation DigiWells: Digital Well Center for Value Creation, Competitiveness and Minimum Environmental Footprint (NFR SFI project no. 309589), funded by Aker BP, ConocoPhillips, Equinor, Harbour Energy, Petrobras, TotalEnergies, Vår Energi, and the Research Council of Norway.

The authors thank ROGII Inc. for providing the academic licenses for Solo Cloud and StarSteer and the relevant training.

\section*{Statement on AI-generated text}
The authors employed OpenAI’s ChatGPT to refine their initial drafts and then carefully revised the AI-generated text to ensure it accurately represented their views and insights.

\appendix
\section{Training Parameters}
\label{app1}
This appendix summarizes in \ref{tab:parameters} the training hyperparameters used across all learning-based experiments to ensure reproducibility and consistency of the reported results.

\begin{table}[ht!]
    \centering
    \caption{Training Parameters}
    \label{tab:parameters}
    \begin{tabular}{|l|c|}
        \hline
        \textbf{Parameter} & \textbf{Value} \\
        \hline
        Number of Episodes & 20,000 \\
        Learning Rate & 0.0005 \\
        Discount Factor (\(\gamma\)) & 0.95 \\
        Batch Size & 64 \\
        Number of Particles in Particle Filter (Training) & 256 \\
        Number of Particles in Particle Filter (Testing) & 2064 \\
        Replay Buffer Size & 50,000 \\
        Episodes before replacements & 1000 \\
        Epsilon Decay Rate & 0.995 \\
        Minimum Epsilon & 0.01 \\
        Target Network Soft Update Rate (\(\tau\)) & 0.005 \\
        Number of Seeds & 10 \\
        \hline
    \end{tabular}
\end{table}

\section{Fine-Scale Trajectory Behavior Near Decision Points}
\label{app2}
\begin{figure}[ht]
    \centering
    \includegraphics[width=0.85\linewidth]{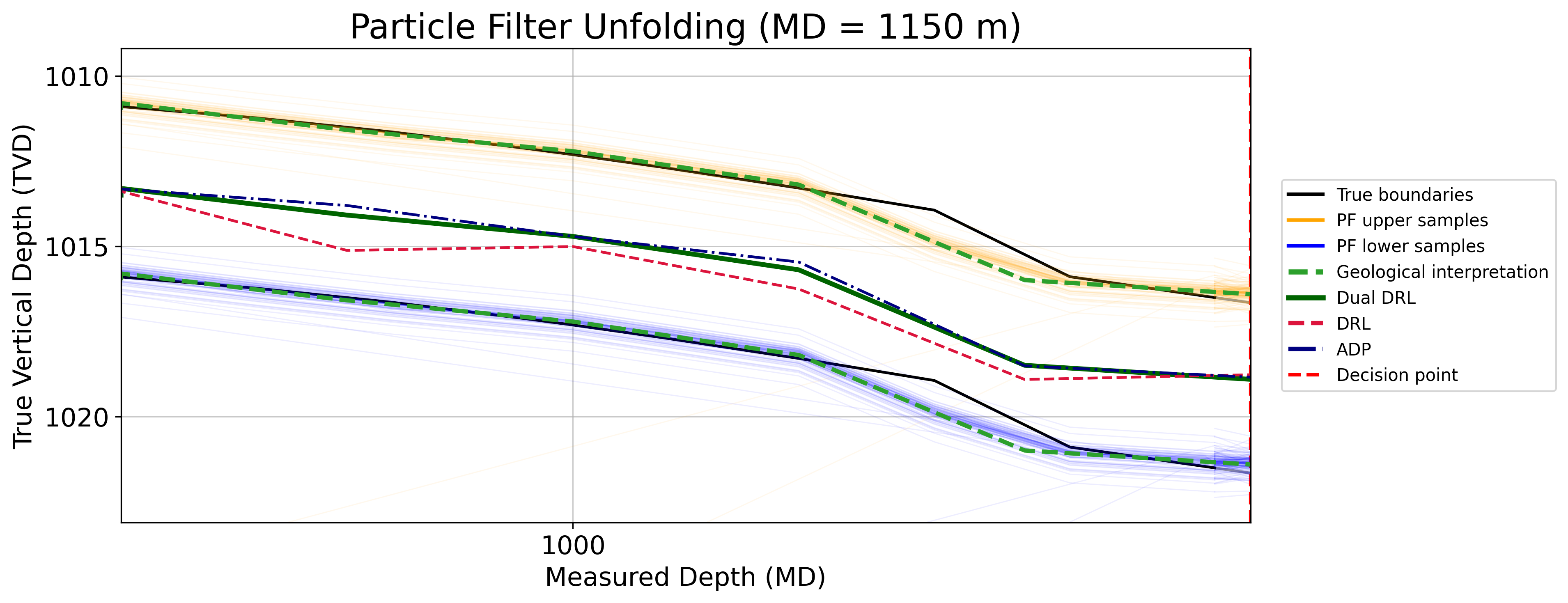}
    \includegraphics[width=0.85\linewidth]{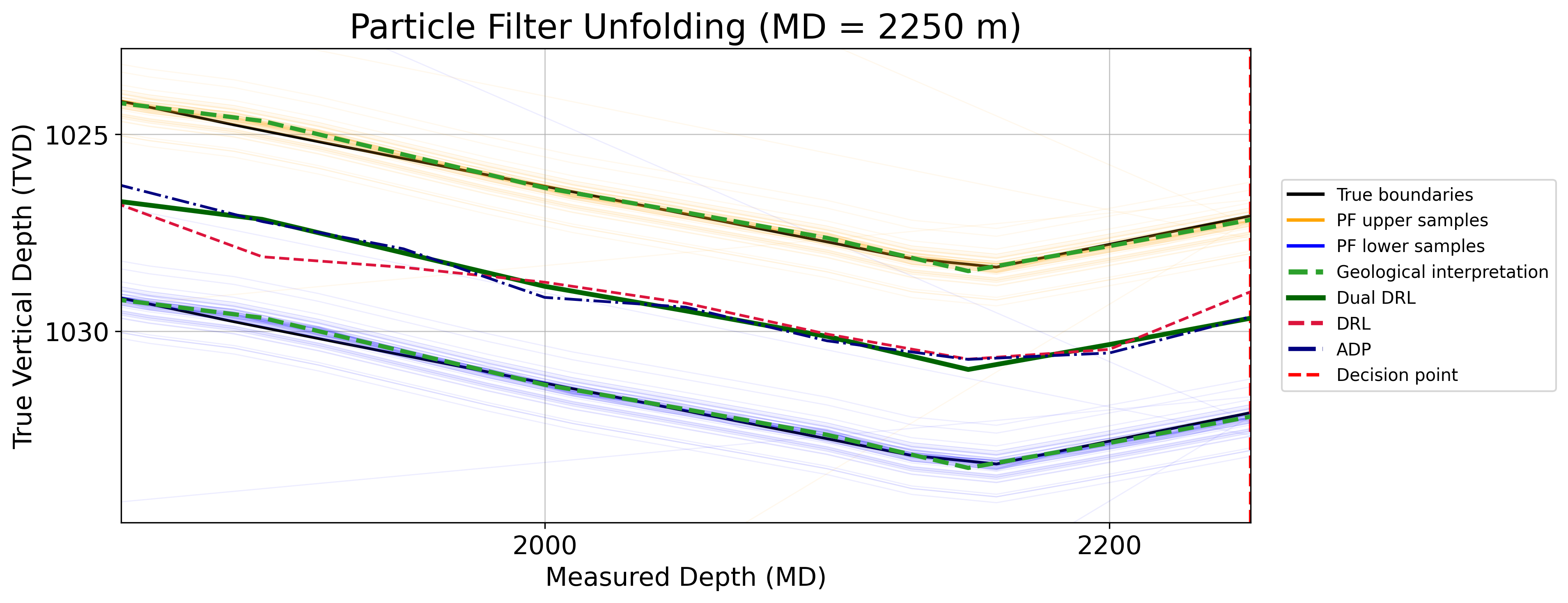}
    \includegraphics[width=0.85\linewidth]{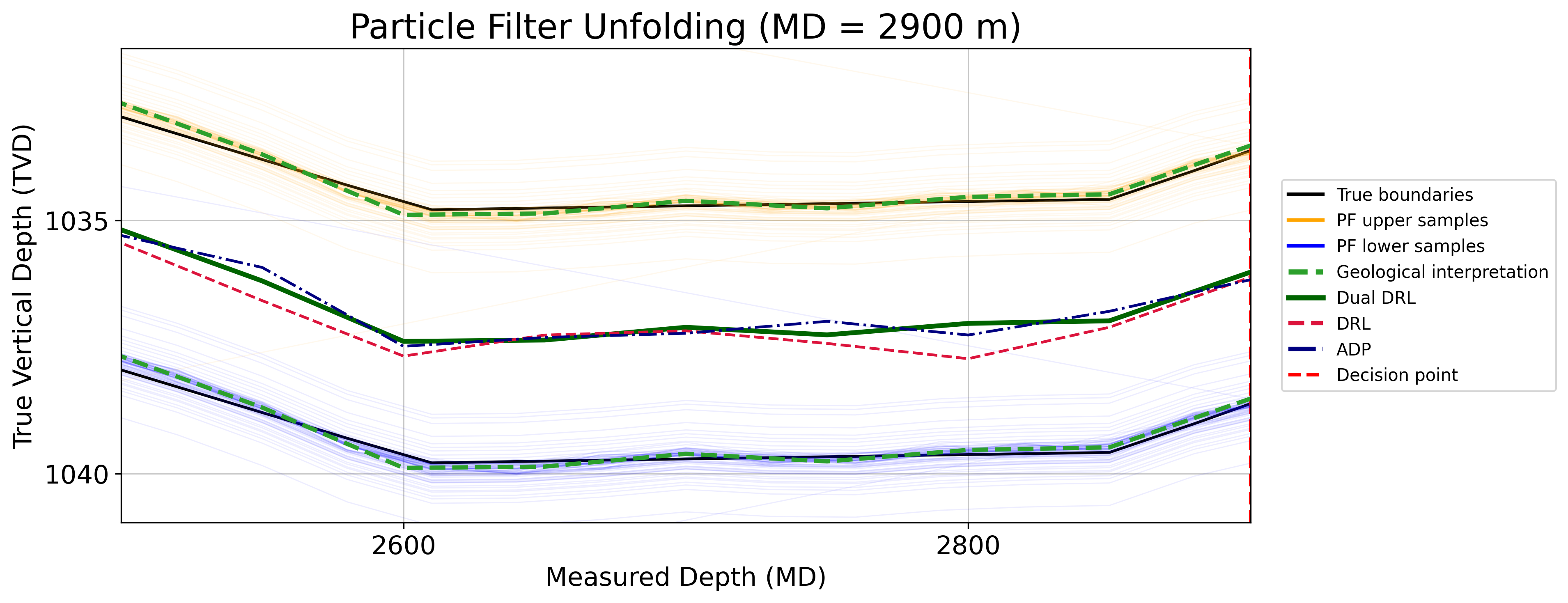}
    \includegraphics[width=0.85\linewidth]{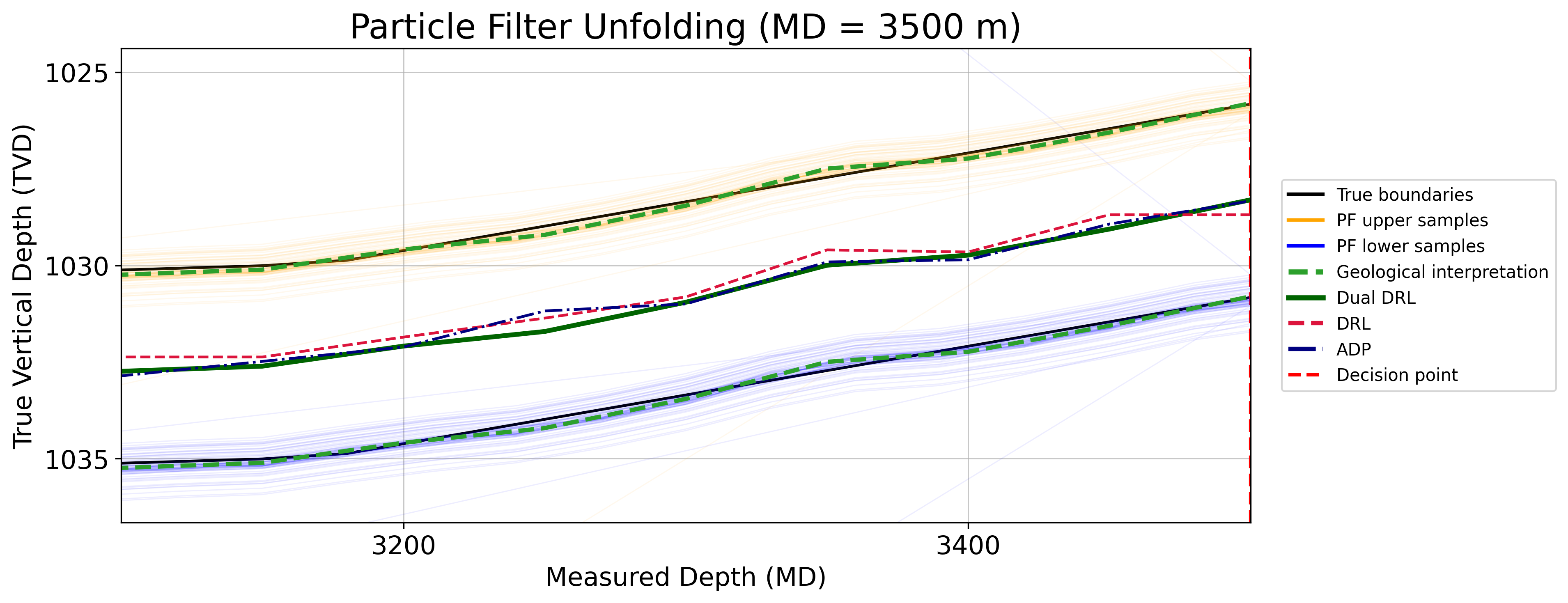}
    \caption{Local trajectory behavior behind the decision point, illustrated over a 400 m window preceding each decision.}
    \label{fig:pf_unfolding_sequence_400}
\end{figure}

This appendix provides supplementary zoomed-in visualizations (\ref{fig:pf_unfolding_sequence_400}) of trajectory behavior behind decision points. 
For each decision step, the last 400 m of the trajectory is shown to illustrate local trajectory evolution and policy behavior at the decision scale. 
These figures complement the global results presented in the main text and are intended to aid interpretation.

\newpage
\bibliographystyle{elsarticle-num} 
\bibliography{references}

@article{10.2118/167433-PA,
    author = {Kullawan, K.  and Bratvold, R.  and Bickel, J.E.},
    title = {A Decision Analytic Approach to Geosteering Operations},
    journal = {SPE Drilling \& Completion},
    volume = {29},
    year = {2014},
    month = {03},
    issn = {1064-6671},
    doi = {10.2118/167433-PA},
}

@article{Alyaev_2019,
   title={A decision support system for multi-target geosteering},
   volume={183},
   ISSN={0920-4105},
   url={http://dx.doi.org/10.1016/j.petrol.2019.106381},
   DOI={10.1016/j.petrol.2019.106381},
   journal={Journal of Petroleum Science and Engineering},
   publisher={Elsevier BV},
   author={Alyaev, Sergey and Suter, Erich and Bratvold, Reider Brumer and Hong, Aojie and Luo, Xiaodong and Fossum, Kristian},
   year={2019},
   month=dec, pages={106381} }

@inproceedings{autosteering,
  title={Automated geosteering while drilling using machine learning. case studies},
  author={Denisenko, Ivan Denisenko and Kuvaev, Igor Andreevich and Uvarov, Igor Borisovich and Kushmantzev, Oleg Evgenievich and Toporov, Artem Igorevich},
  booktitle={SPE Russian Petroleum Technology Conference?},
  pages={D023S009R004},
  year={2020},
  organization={SPE}
}

@inproceedings{alyaev2024distinguish,
  title={DISTINGUISH Workflow: a New Paradigm of Dynamic Well Placement Using Generative Machine Learning},
  author={Alyaev, S and Fossum, K and Djecta, HE and Tveranger, J and Elsheikh, A},
  booktitle={ECMOR 2024},
  volume={2024},
 
  pages={1--16},
  year={2024},
  organization={European Association of Geoscientists \& Engineers}
}

@manual{SoloAPI,
  title        = {Solo REST API Documentation},
  author       = {{Rogii Inc.}},
  year         = {2025},
  url          = {https://api.solo.cloud/},
  note         = {Accessed: 2025-02-11}
}

@misc{high,
      title={High-Precision Geosteering via Reinforcement Learning and Particle Filters}, 
      author={Ressi Bonti Muhammad and Apoorv Srivastava and Sergey Alyaev and Reidar Brumer Bratvold and Daniel M. Tartakovsky},
      year={2024},
      eprint={2402.06377},
      archivePrefix={arXiv},
      primaryClass={cs.LG},
      url={https://arxiv.org/abs/2402.06377}, 
}

@misc{optimal,
    title = {Optimal sequential decision-making in geosteering: A reinforcement learning approach},
    journal = {Geoenergy Science and Engineering},
    volume = {258},
    pages = {214304},
    year = {2026},
    issn = {2949-8910},
    doi = {https://doi.org/10.1016/j.geoen.2025.214304},
}

@article{geo,
    author = {Muhammad, Ressi B. and Cheraghi, Yasaman and Alyaev, Sergey and Srivastava, Apoorv and Bratvold, Reidar B.},
    title = {Geosteering Robot Powered by Multiple Probabilistic Interpretation and Artificial Intelligence: Benchmarking Against Human Experts},
    journal = {SPE Journal},
    pages = {1-15},
    year = {2025},
    month = {01},
    issn = {1086-055X},
    doi = {10.2118/218444-PA},
    url = {https://doi.org/10.2118/218444-PA},
    eprint = {https://onepetro.org/SJ/article-pdf/doi/10.2118/218444-PA/4407193/spe-218444-pa.pdf},
}

@misc{dual,
      title={Dueling Network Architectures for Deep Reinforcement Learning}, 
      author={Ziyu Wang and Tom Schaul and Matteo Hessel and Hado van Hasselt and Marc Lanctot and Nando de Freitas},
      year={2016},
      eprint={1511.06581},
      archivePrefix={arXiv},
      primaryClass={cs.LG},
      url={https://arxiv.org/abs/1511.06581}, 
}

@INPROCEEDINGS{6714080,
  author={Djurić, Petar M. and Bugallo, Mónica F.},
  booktitle={2013 5th IEEE International Workshop on Computational Advances in Multi-Sensor Adaptive Processing (CAMSAP)}, 
  title={Particle filtering for high-dimensional systems}, 
  year={2013},
  volume={},
  number={},
  pages={352-355},
  doi={10.1109/CAMSAP.2013.6714080}}

@inproceedings{NIPS2000_e0ab531e,
 author = {Shelton, Christian},
 booktitle = {Advances in Neural Information Processing Systems},
 editor = {T. Leen and T. Dietterich and V. Tresp},
 pages = {12},
 publisher = {MIT Press},
 title = {Balancing Multiple Sources of Reward in Reinforcement Learning},
 volume = {13},
 year = {2000}
}

@article{KULLAWAN201890,
title = {Sequential geosteering decisions for optimization of real-time well placement},
journal = {Journal of Petroleum Science and Engineering},
author = {K. Kullawan and R.B. Bratvold and J.E. Bickel},
volume = {165},
pages = {90-104},
year = {2018},
issn = {0920-4105},

}

@book{Sutton1998,
  author = {Sutton, Richard S. and Barto, Andrew G.},
  edition = {Second},
  publisher = {The MIT Press},
  title = {Reinforcement Learning: An Introduction},
  url = {http://incompleteideas.net/book/the-book-2nd.html},
  year = {2018 }
}

@InProceedings{10.1007/978-3-031-97554-7_14,
author="Djecta, Hibat Errahmen
and Alyaev, Sergey
and Fossum, Kristian
and B. Bratvold, Reidar
and Muhammad, Ressi Bonti
and Srivastava, Apoorv",
editor="Paszynski, Maciej
and Barnard, Amanda S.
and Zhang, Yongjie Jessica",
title="Uncertainty-Aware Well Placement: Simulator-Verified Dual-Network Reinforcement Learning Approach Meets Particle Filters",
booktitle="Computational Science -- ICCS 2025 Workshops",
year="2025",
publisher="Springer Nature Switzerland",
address="Cham",
pages="188--202",
isbn="978-3-031-97554-7"
}

@article{Evensen2003,
  author  = {Evensen, Geir},
  title   = {The Ensemble Kalman Filter: theoretical formulation and practical implementation},
  journal = {Ocean Dynamics},
  year    = {2003},
  volume  = {53},
  number  = {4},
  pages   = {343--367},
  doi     = {10.1007/s10236-003-0036-9},
  url     = {https://doi.org/10.1007/s10236-003-0036-9},
  issn    = {1616-7228}
}

@misc{chen2017tutorialkerneldensityestimation,
      title={A Tutorial on Kernel Density Estimation and Recent Advances}, 
      author={Yen-Chi Chen},
      year={2017},
      eprint={1704.03924},
      archivePrefix={arXiv},
      primaryClass={stat.ME},
      url={https://arxiv.org/abs/1704.03924}, 
}

@article{DBLP:journals/corr/MnihKSGAWR13,
  author       = {Volodymyr Mnih and
                  Koray Kavukcuoglu and
                  David Silver and
                  Alex Graves and
                  Ioannis Antonoglou and
                  Daan Wierstra and
                  Martin A. Riedmiller},
  title        = {Playing Atari with Deep Reinforcement Learning},
  journal      = {CoRR},
  volume       = {abs/1312.5602},
  year         = {2013},
  url          = {http://arxiv.org/abs/1312.5602},
  eprinttype    = {arXiv},
  eprint       = {1312.5602},
  bibsource    = {dblp computer science bibliography, https://dblp.org}
}

@inproceedings{
djecta2025geosteering,
title={Geosteering Through the Lens of Decision Transformers: Toward Embodied Sequence Decision-Making},
author={Hibat Errahmen DJECTA and Sergey Alyaev and Kristian Fossum and Reidar B. Bratvold and Dan Sui},
booktitle={NeurIPS 2025 Workshop on Embodied World Models for Decision Making},
year={2025},
pages = {12},
url={https://openreview.net/forum?id=QXLWeLJ0ub}
}

@misc{chen2021decisiontransformerreinforcementlearning,
      title={Decision Transformer: Reinforcement Learning via Sequence Modeling}, 
      author={Lili Chen and Kevin Lu and Aravind Rajeswaran and Kimin Lee and Aditya Grover and Michael Laskin and Pieter Abbeel and Aravind Srinivas and Igor Mordatch},
      year={2021},
      eprint={2106.01345},
      archivePrefix={arXiv},
      primaryClass={cs.LG},
      url={https://arxiv.org/abs/2106.01345}, 
}

@article{6ce60d164cfb4c5cb00bfc2a662a8e4d,
title = "Development of a nonlinear predictive controller for mitigation of motion sickness in autonomous vehicles through multi-objective control of lateral and roll dynamics",
author = "Arslan, \{M. Selcuk\} and Ibrahim Kucukdemiral and Farrag, \{Mohamed E.\}",
year = "2025",
month = mar,
doi = "10.1016/j.rineng.2024.103816",
language = "English",
volume = "25",
journal = "Results in Engineering",
issn = "2590-1230",
publisher = "Elsevier B.V.",

}

@article{veettil2020bayesian,
  title   = {Bayesian Geosteering Using Sequential Monte Carlo Methods},
  author  = {Veettil, D. R. A. and Clark, K.},
  journal = {Petrophysics},
  volume  = {61},
  number  = {1},
  pages   = {99--111},
  year    = {2020},
  doi     = {10.30632/PJV61N1-2020a4}
}

\end{document}